# Measure Selection: Notions of Rationality and Representation Independence

**Manfred Jaeger**
Max-Planck-Institut für Informatik
Im Stadtwald, 66123 Saarbrücken, Germany
jaeger@mpi-sb.mpg.de

## Abstract

We take another look at the general problem of selecting a preferred probability measure among those that comply with some given constraints. The dominant role that entropy maximization has obtained in this context is questioned by arguing that the minimum information principle on which it is based could be supplanted by an at least as plausible "likelihood of evidence" principle. We then review a method for turning given selection functions into representation independent variants, and discuss the tradeoffs involved in this transformation.

## 1 INTRODUCTION

An ever recurring theme in probabilistic inference is the selection of preferred probability measures from some set of possible choices: we are given a state space $A$, a subset $J$ of the set $\Delta A$ of probability measures on $A$, and are asked to identify a set $I(J) \subseteq J$ of measures that fulfill certain desiderata.

One example that instantiates this abstract schema is the selection of a prior probability distribution in Bayesian statistics. In this case, $J$ is the set of all probability measures (usually restricted to a suitable parametric family) consistent with our prior information, and $I(J)$ is the element selected as our prior, usually on the basis of some minimum information principle. Another example is default semantics for probabilistic knowledge bases. In this case $A$ is the set of all models of some propositional language, $J \subseteq \Delta A$ is the set of models of some knowledge base $KB$ in some probabilistic extension of propositional logic, and $I(J)$ is a set of default models of $KB$ that reflect certain commonsense inferences to be drawn from $KB$.

A number of studies (Shore & Johnson 1980, Paris & Vencovská 1990, Paris 1994) have addressed the question of what general *rationality principles* should guide our choice of $I(J)$, and which formal method can be used to implement these principles. From these considerations, entropy maximization emerges as the unique selection rule that satisfies our needs in general. Another selection rule, called center of mass by Paris (1994), against this background, does not seem to be a serious competitor of entropy maximization.

The purpose of the present paper is twofold. In the first part it is argued that center of mass inference, in spite of its poor performance with respect to theoretical rationality principles, is deeply entrenched in commonsense probabilistic reasoning. As an explanation for this phenomenon we argue that whereas entropy maximization is derived from a general *minimum information principle*, center of mass inference might be justified by a no less viable *likelihood of evidence* principle. In the second part of the paper we discuss the problem of representation dependence, both of center of mass and of maximum entropy inference. In (Jaeger 1996) a method was proposed that transforms selection rules $I$ into variants $\tilde{I}$ that are representation independent. In the current paper, results are presented that show what useful properties of $I$ (particularly as expressed by rationality principles) we have to trade in for representation independence, and which of these properties will be preserved in $\tilde{I}$.

## 2 PRINCIPLES OF MEASURE SELECTION

In this section the basic definitions for measure selection functions and rationality principles are provided. We here present a purely semantic set of definitions. This is to say, we introduce the selection function $I$ as operating on sets of probability measures $\Delta A$, and that we formulate rationality principles as conditions on the geometric form of $I(J)$ given the geometric form of $J$. The alternative approach is syntactic: in that approach one focuses on a specific probabilistic representation language, the class $A$ of its models, and sets $J \subseteq \Delta A$ that are described by knowledge bases in the language. The selection function $I$ is then seen as operating on knowledge bases $KB$, and rationality principles im-pose conditions on the syntactic form of statements valid in $I(KB)$ in terms of the syntactic form of $KB$. This approach is taken by Paris (1994).

For a finite state space $A$ of size $n$, after ordering the elements of $A$ in an arbitrary way, we can identify $\Delta A$ with

$$\Delta^n := \{(x_1, \ldots, x_n) \in \mathbf{R}^n \mid x_i \geq 0, \ \sum x_i = 1\}.$$

A *measure selection function* $I$ is any function that for every $n \in \mathbf{N}$ maps $\mathscr{P}(\Delta^n)$ (the powerset of $\Delta^n$) into $\mathscr{P}(\Delta^n)$, such that $I(J) \subseteq J$ for all $J \subseteq \Delta^n$, and $I(\pi(J)) = \pi(I(J))$ for all permutations $\pi$ of $(1, \ldots, n)$. This last condition makes sure that a measure selection function defined on $\{\Delta^n \mid n \in \mathbf{N}\}$ can be applied unambiguously to $\Delta A$ for arbitrary finite $A$, because it then does not matter what particular order we use on $A$. Note that in contrast with Shore and Johnson (1980) and Paris and Vencovská (1990) we do not demand $I$ to be point-valued, i.e. $I(J)$ to be a single measure in $\Delta^n$.

In this paper we are mostly concerned with two particular selection functions. The first is entropy maximization, denoted $I_{me}$, where from $J \subseteq \Delta^n$ we select those $P \in J$, $P = (p_1, \ldots, p_n)$, for which $H(P) := -\sum_{i=1}^{n} p_i ln(p_i)$ is maximal. $I_{me}(J)$ is a singleton for closed and convex $J$; it is nonempty for closed $J$. When $J$ is not closed, there need not exist an entropy maximal element in $J$, in which case $I(J) = \emptyset$.

The second selection function we here consider is $I_{cm}$, the center of mass selection function. The center of mass $(\bar{p}_1, \ldots, \bar{p}_n)$ of $J \subseteq \Delta^n$ is determined by

$$\bar{p}_i := lim_{\epsilon \to 0} \frac{1}{\int_{J(\epsilon)} d\lambda^n} \int_{J(\epsilon)} p_i d\lambda^n, \tag{1}$$

where $J(\epsilon)$ is the "$\epsilon$-hull" around $J$, i.e. the set containing all points of $\mathbf{R}^n$ with a Euclidean distance smaller than $\epsilon$ to some element of $J$; $\lambda^n$ is the $n$-dimensional Lebesgue measure. Taking the limit over the $J(\epsilon)$ in (1) is necessary, because $J \subseteq \Delta^n$ has Lebesgue measure zero, so that both integrals in (1) would be zero when taken over $J$. $I_{cm}(J)$ is either a singleton (when the limit (1) exists, and the center of mass so defined lies inside $J$ – this is guaranteed for convex $J$), or else empty.

Next, we briefly formulate the most important formal conditions for $I$ that were introduced as *consistency axioms* by Shore and Johnson (1980), and as *(rationality) principles* by Paris and Vencovská (1990). We state these conditions in a form that is semantic and generalizes the previous versions (given for point-valued selection rules $I$) to set-valued $I$. For intuitive motivations of the principles the reader is referred to (Shore & Johnson 1980) and (Paris 1994). Using the terminology of Paris and Vencovská, we here consider the principles of relativization, obstinacy, independence, and irrelevant information.

To define the *relativization principle* we need the following notation: if $A$ is a state space, $B \subseteq A$, and $P \in \Delta A$, then $P|B \in \Delta B$ denotes the conditional distribution of $P$ on $B$. The notation $B^c$ is used for the complement of $B$. $I$ satisfies the *relativization principle*, if the following holds: whenever $J$ is of the form

$$J = \{P \in \Delta A \mid P|B \in J^B, \ P|B^c \in J^{B^c}\} \tag{2}$$

for some $J^B \in \Delta B$, $J^{B^c} \in \Delta B^c$ with $I(J^{B^c}) \neq \emptyset$, then

$$I(J)|B := \{P|B \mid P \in I(J), P(B) > 0\} = I(J^B).$$

The *obstinacy principle* says that whenever $I(J) \subseteq J^* \subseteq J$, then $I(J^*) = I(J)$.

The *independence principle* does not generalize as unambiguously from point-valued $I$ to set-valued $I$, as the previous two principles. We therefore here introduce two variants, strong independence and weak independence. To define these principles, let $A = B \times C$ be a product space; for $P \in \Delta A$ let $P{\restriction}B$, $P{\restriction}C$ denote the marginal distributions induced by $P$ on $B$ and $C$, respectively. Consider $J \in \Delta A$ of the form

$$J = \{P \mid P{\restriction}B \in J^B, \ P{\restriction}C \in J^C\} \tag{3}$$

for some $J^B \subseteq \Delta B$, $J^C \subseteq \Delta C$. We say that $I$ satisfies the *strong independence principle* if for such $J$ we have

$$\begin{aligned} I(J) &= I(J^B) \otimes I(J^C) \tag{4} \\ &:= \{P \otimes Q \mid P \in I(J^B), Q \in I(J^Q)\}, \end{aligned}$$

where $\otimes$ is the standard product of measures.

Conditions (4) is very strong, as it encodes an independence assumption not expressible by linear constraints, and that, in general, can be satisfied only by selecting non-convex $I(J)$, even for convex $J$. A "linear approximation" to (4) is weak independence: we say that $I$ satisfies the *weak independence principle* if for $J$ of the form (3), and all $B' \subseteq B$, $C' \subseteq C$ we have

$$\begin{aligned} inf\, I(J)(B' \times C') &= inf\, I(J^B)(B') \cdot inf\, I(J^C)(C') \\ sup\, I(J)(B' \times C') &= sup\, I(J^B)(B') \cdot sup\, I(J^C)(C') \end{aligned}$$

where, e.g. $inf\, I(J)(B' \times C') := inf\{P(B' \times C') \mid P \in I(J)\}$.

Closely related to the independence principles is the *irrelevant information principle*: this principle is satisfied by $I$ if for $J$ of the form (3) we get

$$I(J){\restriction}B = I(J^B). \tag{5}$$

To these familiar rationality principles, we add one more technical property that will be needed below: $I$ is called *dimension independent* if $I(J \times \{0\}) = I(J) \times \{0\}$.

| | Relativization | Obstinacy | Strong Independence | Weak Independence | Irrelevant Information |
|---|---|---|---|---|---|
| $I_{me}$ | + | + | + | + | + |
| $I_{cm}$ | + | - | - | - | - |
| $I \to \bar{I}$ preserves | + | - | - | + | +<br>I dim.indep. |
| $\bar{I}_{me}$ | + | - | - | + | + |
| $\bar{I}_{cm}$ | + | - | - | - | - |

Table 1: Rationality Principles for Some Selection Functions

## 3 CENTER OF MASS VS. MAXIMUM ENTROPY

Table 1 lists a number of results on which selection functions satisfy the rationality principles listed in the previous section. For the time being, we are only concerned with the first two lines in the table, which list (for the most part well known) results about maximum entropy and center of mass. $I_{me}$, of course, satisfies all of the listed principles, and the fact that (under some mild additional assumptions) it is the only selection function that will achieve this has been proposed as a reason to prefer entropy maximization over all other selection rules (Shore & Johnson 1980, Paris & Vencovská 1990). Paris (1994) is careful to point out that such a justification of entropy maximization hinges on what he calls *Watts assumption*: the set $J$ of possible measures encodes everything that is known to the expert whose judgments the selection function is supposed to model. Given that $J$ is all we know, and that nonetheless we are forced to choose some smaller (or unique) $I(J) \subseteq J$, it then is natural to let this choice be guided by the

> *Minimal information desideratum (MID): through the selection of $I(J)$ as little as possible additional information beyond $J$ should be assumed.*

The rationality principles now simply are rigorous conditions capturing several aspects of the *MID*.

Even though the *MID* is rather plausible, and persuasively leads us to the maximum entropy principle, it is not hard to find evidence that center of mass inference has an intuitive appeal as a model for commonsense probabilistic inference that is at least as great as that of maximum entropy. Recent works in which center of mass inference is applied in some guise are (Druzdzel & van der Gaag 1995) and (Grove & Halpern 1997). One source from which more or less explicit instances of center of mass inference originate is the fact that this selection rule arises naturally from an analysis of our selection problem from the point of view of Bayesian statistics. Let us briefly turn to this connection.

In Bayesian statistics a belief state over a state space $A$ is maintained by a probability distribution $P^\Theta$ on some pa-rameter space $\Theta$. Each $\theta \in \Theta$ determines a distribution $P_\theta$ on $A$, so that a distribution $P^A$ on $A$ is given by

$$P^A(B) = \int_\Theta P_\theta(B) dP^\Theta. \tag{6}$$

The distribution defined by (6) is known as the *predictive distribution*. The probability distribution $P^\Theta$ is updated on observations of elements of $A$ according to Bayes' rule.

At first sight, this Bayesian procedure seems to have a different field of application than measure selection rules like maximum entropy, because it takes as input an observed event, rather than a set of admissible probability distributions. It is well known, however, that by suitable modeling, Bayesian conditioning and entropy maximization can be made to bear on the same problems, and that they then often yield incompatible results (see (Uffink 1996) for a review of results).

One perspective we can adopt in order to process information presented as admissible subsets $J \subseteq \Delta A$ of distributions within the Bayesian framework is to view $J$ as an observed event $J^\Theta := \{\theta \in \Theta \mid P_\theta \in J\}$ in $\Theta$. We can then simply condition $P^\Theta$ on $J^\Theta$, and obtain via the posterior distribution $\bar{P}^\Theta$ a new predictive distribution on $A$.

In general, $\Theta$ will not parameterize the whole set $\Delta A$, i.e. there are $P \in \Delta A$ with $P \neq P_\theta$ for all $\theta$. For the case of finite $A$ that we are here concerned with, however, we can choose $\Theta$ so that $\{P_\theta \mid \theta \in \Theta\} = \Delta A$. A canonical choice is $\Theta = \Delta^n$, for which we then get the trivial identity $P_\theta = \theta$.

Before the constraints $J$ have been obtained, our distribution $P^\Theta$ on $\Theta$ would be chosen as a non-informative prior, which here is the Lebesgue measure on $\Delta^n$ (normalized, so as to yield a probability measure). The posterior $\bar{P}^\Theta$, after conditioning on $J$ (or on $J(\epsilon)$ ($\epsilon \to 0$), if $J$ has measure 0), then is simply the Lebesgue measure restricted to $J$, and the new predictive distribution on $A$ is given by (1). Thus, we have seen that center of mass inference really can be understood as an instance of Bayesian conditioning and marginalization. Since these are inference techniques that we would certainly not consider irrational, and yet center of mass fails the rationality principles emanating from the *MID*, we have to question the *MID* as the exclusive notion of rationality, and look for alternatives. We motivate a proposal for a dif-ferent principle on which to base our notion of rationality by two very simple examples.

First, consider the scenario where we are told that a certain coin is biased, and yields heads with probability in the interval $[0.6, 0.9]$. What should our assumption be on the exact value of $P(heads)$? Maximum entropy prescribes $P(heads) = 0.6$. How can we then justify the (arguably more intuitive) center of mass solution $P(heads) = 0.75$? In this case, we might reason as follows. We will first construct a likely scenario for how the original information $P(heads) \in [0.6, 0.9]$ was obtained. The most plausible such scenario here is to assume that the coin in question has been tossed a number of times, that the relative frequency of heads has been observed, and then has been imbedded in the confidence interval $[0.6, 0.9]$ chosen wide enough to render the statement $P(heads) \in [0.6, 0.9]$ a virtual certainty. Under these assumptions, it would be most reasonable to take for our inferred value of $P(heads)$ the originally observed frequency, which would be assumed to have been 0.75.

In a second example, suppose that we are told that in the first democratic parliamentary elections in the newly independent Republic of Transcaucasia the Progressive Democratic Party (PDP) has gained at least 5% of the votes, the National Unity Party (NUP) has gained at least 55% of the votes, and that these were the only two parties on the ballot. What should our belief be about the actual result of the election, i.e. what is our estimate of the probability $P(\text{PDP})$ of a random voter having voted for PDP? Maximum entropy says $P(\text{PDP}) = 0.45$, but a probably much more sensible estimate is gained when we assume that the given constraint $P(\text{PDP}) \in [0.05, 0.45]$ reflects the intermediate result after 60% of the votes have been counted, and that the final result is obtained by extrapolating this partial count to all votes cast, which would yield approximately 8% for PDP, i.e. $P(\text{PDP}) = 0.08$. This is different from the center of mass solution $P(\text{PDP}) = 0.25$, but shares with it the important characteristic of choosing a value in the interior of the constraint set, rather than on the boundary, as maximum entropy will do.

The two examples illustrate a form of probabilistic inference not accounted for by entropy maximization: when we are given the information that the "true" distribution $P$ belongs to $J$, we are not limited to take this information at its face value only, i.e. as a constraint on which $P \in \Delta^n$ we may choose, but we can also take advantage of the "meta-information" that "$P \in J$" is exactly what we got to know. In the examples above we have used this meta-information together with plausible assumptions on how information $J$ typically is generated to arrive at our results. More formally, $J$ has been interpreted as the value of a random variable whose distribution is determined by the "true" value $P \in \Delta^n$ we want to infer. Thus, the selection procedure $J \to I(J)$ essentially becomes a statistical parameter estimation problem, and the guiding principle for this selection

can be formulated as the

> *Maximum likelihood of evidence desideratum (MLED): the set $I(J)$ should contain those distributions that are most likely to produce the information $J$.*

One might object that in the discussion of our examples we have blatantly violated Watts assumption, because our arguments made use of the semantic content of the variables *heads* and PDP, which is information not contained in $J$. This is true as far as the two specific results argued for in the examples are concerned. It does not, however, compromise the MLED as a general principle that we can aim for even when such semantic background information is missing. The question now, of course, is how the philosophical MLED can be sharpened into formal conditions in the same way that the rationality principles sharpen the MID, and what, if any, selection functions satisfy these conditions. A first such condition that one might consider is to require that for convex $J$ the selected set $I(J)$ lies in the (relative) interior of $J$. This very weak condition already eliminates maximum entropy, but retains center of mass as a possible choice.

## 4    REPRESENTATION INDEPENDENCE

### 4.1    THE ISSUE

In the previous section it was argued that the failure of the rationality principle for center of mass need not be construed as a conclusive argument against this selection rule. There is one rationality principle, however, whose violation by center of mass really is quite disturbing: center of mass is not *language invariant* (Paris 1994). This means that the result obtained by applying $I_{cm}$ to some knowledge base *KB* depends on our assumptions of what additional propositional variables there exist in our probability space except those actually mentioned in *KB*. If, for instance, in the coin-tossing example of section 3 we had assumed that besides the variables *heads* there also is a variable *quarter* in our vocabulary (standing for the fact that the coin in question is a quarter), then the mere presence of this additional variable will change our results for $P(heads)$, even though there is no information given about *quarter*, let alone any information linking *heads* to *quarter*.

Maximum entropy satisfies language independence, but it fails to satisfy another property that can be seen as a further rationality principle: representation independence. As a very simple illustration of the problem, we may compare the results obtained by applying entropy maximization to the two knowledge bases $KB := P(A) \leq 0.9$ and $KB_1 := P(A_1 \vee A_2) \leq 0.9$, where A, $A_1$, $A_2$ are propositional variables. In the first case entropy maximazation yields $P(A) = 0.5$; in the second case $P(A_1 \vee$ $A_2$) = 0.75. That this is an often undesirable behavior becomes clear when we substitute e.g. $A \equiv$ "there exists life on Mars", $A_1 \equiv$ "there exists plant life on Mars", $A_2 \equiv$ "there exists animal life on Mars". Based on examples like these, we can roughly describe representation dependence of entropy maximization (and other selection functions) as the property of returning results that depend on non-essential choices of language and syntax, rather than on semantic content only. It was not until recently, that informal, example-based, descriptions of representation dependence received more rigorous underpinnings. Paris (1994) appears to have been the first to supply a precise concept of representation independence by introducing his *atomicity principle*. This principle requires that inferred probability values do not change when a knowledge base is translated by replacing a propositional variable by a formula in a new language, just as A was replaced by $A_1 \vee A_2$ in our example above. Paris then shows that this principle can not be satisfied by a selection function $I$ that yields unique values $I(J)$ for closed and convex $J$. Paris and Vencovská (1997) later argue that atomicity is not a reasonable principle in the first place, because the fact whether representations A or $A_1 \vee A_2$ are used is relevant information that may very well influence our inferences.

Halpern and Koller (1995) give a semantic definition of representation independence, based on *embeddings* $f : A \rightarrow B$ of state spaces. This definition subsumes the atomicity principle, but can also be applied to selection problems not framed within the context of a propositional probabilistic logic.

In (Jaeger 1996) a generalization of atomicity along a different line is provided by developing a concept of representation independence for arbitrary nonmonotonic logics. That concept, like atomicitiy, is syntactic, based on *interpretations* between formal languages.

In the remainder of this section, a definition of representation independence is given that combines elements from the ones found in (Halpern & Koller 1995) and (Jaeger 1996). In order to stay in line with the other definitions used in the present paper, the definitions we provide are semantic, but our motivation for these definitions very much derives from syntactic considerations.

A definition of representation independence essentially hinges on a definition of what constitutes a representation change. In our first example, a representation change was given by the syntactic interpretation $f : A \mapsto A_1 \vee A_2$, which induces the mapping $f : KB \mapsto KB_1$. Here the representation $KB_1$ is a refinement of the representation $KB$, obtained by moving from a simpler to a richer language. In general, we will also want to deal with alternative representations of the same information, none of which is a strict refinement of the other. For instance, let $KB_2 = P(B_1 \wedge B_2) \leq 0.9$. $KB_1$ and $KB_2$ now represent the same informa-tion with respect to the correspondence $A_1 \vee A_2 \leftrightarrow B_1 \wedge B_2$. It is convenient to model such a correspondence as mediated by a third "common ground" language, so that both representations are interpretations from this (poorer) language. Here we can choose $\{A\}$ as the common ground language, and obtain $KB_2 = g(KB)$ under the interpretation $g : A \mapsto B_1 \wedge B_2$. In the terminology of (Jaeger 1996), $KB_1$ and $KB_2$ would be called *representational variants* with respect to $f$ and $g$.

The semantic analogues of syntactic interpretations are *embeddings* of state spaces (Halpern & Koller 1995).

**Definition 4.1** Let $A, B$ be finite state spaces. An embedding of $A$ in $B$ is any function $f : A \rightarrow \mathscr{P}(B)$ with $a_1 \neq a_2 \Rightarrow f(a_1) \cap f(a_2) = \emptyset$, and $B = \cup_{a \in A} f(a)$.

In the case where $A$ and $B$ are the sets of models of propositional languages $L_A$ and $L_B$, any syntactic interpretation $f : X \mapsto \phi_X$ ($X$ ranging over the propositional variables of $L_A$, $\phi_X$ being a formula in $L_B$) induces an embedding of $A$ in $B$: a model $a$ for $L_A$ is mapped to the set of models $f(a) \subseteq B$ (possibly empty) in which the formulas $\phi_X$ have the same truth values as the variables $X$ have in $a$.

Embeddings $f : A \rightarrow B$ induce a mapping $f : \Delta A \rightarrow \mathscr{P}(\Delta B)$ via $f(P) = \{Q \in \Delta B \mid \forall a \in A : Q(f(a)) = P(a)\}$. Note that we get $f(P) = \emptyset$ exactly when there exists $a \in A$ with $P(a) > 0$ and $f(a) = \emptyset$. For $J \subseteq \Delta A$ we write $f(J)$ for $\cup_{P \in J} f(P)$. For $Q \in \Delta B$ we define $\bar{f}(Q) \in \Delta A$ via $\bar{Q}(a) := Q(f(a))$. Finally, for $H \subseteq \Delta B$, let $\bar{f}(H) := \{\bar{f}(Q) \mid Q \in H\}$.

From our informal discussion of when two knowledge bases are representational variants, and the definition of embeddings, we now derive a formal semantic definition of representational variants.

**Definition 4.2** Let $A, B, C$ be state spaces, $f : C \rightarrow A$, $g : C \rightarrow B$ be embeddings. Let $J \subseteq \Delta A$, $H \subseteq \Delta B$. We say that $J$ and $H$ are representational variants with respect to $f$ and $g$, written $J \xrightarrow{f,g} H$, iff

$$\bar{f}(J) = \bar{g}(H). \tag{7}$$

and

$$J = f(\bar{f}(J)), \qquad H = g(\bar{g}(H)) \tag{8}$$

Condition (7) says that $J$ and $H$ contain the same information about the common ground state space $C$. Condition (8) essentially means that $J$ and $H$ do not contain any additional information about $A$ and $B$, respectively, that is not given as a translated constraint on $C$. Condition (8) is quite restrictive, and in (Jaeger 1996) was not part of the definition of representational variants. We include it here solely for convenience, because our subsequent results only apply to this restricted notion.

**Definition 4.3** A measure selection function $I$ is called representation independent, iff for any $A, B, C, f, g, J, H$ as in definition 4.2, we have that $J \xleftrightarrow{f,g} H$ implies

$$\bar{f}(I(J)) = \bar{g}(I(H)). \tag{9}$$

Definition 4.3 is very similar to the one given by Halpern and Koller (1995), only that their unidirectional notion of representation shifts is replaced by the symmetrical notion of representational variants. The definition of Halpern and Koller is here covered by the special case $B = C$, $g$ the identity function, and $f$ a *faithful* embedding (i.e. $f(c) \neq \emptyset$ for all $c \in C$). Also, it is easy to see that representation independence in the sense of definition 4.3 implies language independence.

## 4.2 REPRESENTATION INDEPENDENT SELECTION FUNCTIONS

In this section we study representation independent selection functions. First, we review a construction presented in (Jaeger 1996) that allows us to transform a given selection function $I$ into a representation independent variant $\tilde{I}$. To motivate this construction, first assume that we are given the situation of definition 4.2, i.e. we have the three state spaces $A, B, C$, the embeddings $f, g$, and the sets $J, H$ with $J \xleftrightarrow{f,g} H$. We want to select subsets $\tilde{I}(J), \tilde{I}(H)$ such that (9) holds. Here there is an obvious way of doing this: we simply use any selection function $I$ to choose $I(\bar{f}(J)) = I(\bar{g}(H))$, and then let $\tilde{I}(J) := f(I(\bar{f}(J))), \tilde{I}(H) := g(I(\bar{g}(H)))$. Because of (8) we have $\tilde{I}(J) \subseteq J, \tilde{I}(H) \subseteq H$.

The problem, of course, is that in general we are not given a special scenario $J \xleftrightarrow{f,g} H$ for which (9) has to be satisfied, but only some $J \subseteq \Delta A$ from which we have to choose $\tilde{I}(J)$ such that (9) holds for every possible instantiation of the schema $J \xleftrightarrow{f,g} H$. The key to defining $\tilde{I}$ is the observation that there is a simplest state space $C$ and an embedding $f : C \rightarrow A$ such that $J = f(\bar{f}(J))$. For any selection function $I$ we can therefore define $\tilde{I}(J) := f(I(\bar{f}(J)))$, which then defines a representation independent selection function $\tilde{I}$. The following definitions and results taken from (Jaeger 1996) (here somewhat reformulated to fit into our semantic framework) describe the construction.

**Theorem 4.4** Let $J \subseteq \Delta A$. There exists a smallest state space $S_J$ and an embedding $f_J : S_J \rightarrow A$, such that $J = f_J(\bar{f}_J(J))$. $S_J$ and $f_J$ are unique up to renaming the elements of $S_J$.

The embedding $f_J$ induces a partition $\{f_J(s) \mid s \in S_J\}$ on $A$. In fact, we can choose this partition itself as a canonical representation of $S_J$. The embedding $f_J$ then simply is the identity. The function $\bar{f}_J : \Delta A \rightarrow \Delta S_J$ becomes the restriction $\bar{f}_J(P) = P \upharpoonright S_J$, and the mapping $f_J : \Delta S_J \rightarrow$

$\Delta A$ induced by $f_J$ is the extension operator:

$$Ext(Q) := f_J(Q) = \{P \in \Delta A \mid P \upharpoonright S_J = Q\}.$$

Thus, theorem 4.4 can be restated as follows: for $J \subseteq \Delta A$ there exists a unique coarsest partition $S_J$ of $A$, s.t.

$$J = Ext(J \upharpoonright S_J). \tag{10}$$

In the following, we will understand $S_J$ to refer to this partition.

**Definition 4.5** Let $I$ be a measure selection function. The measure selection function $\tilde{I}$ is defined by

$$\tilde{I}(J) := Ext(I(J \upharpoonright S_J)). \tag{11}$$

**Theorem 4.6** When $I$ is a dimension independent selection function, then $\tilde{I}$ is representation independent.

The following short example illustrates the definitions and theorems formulated so far.

**Example 4.7** Let $A = \{a_1, a_2, a_3, a_4\}, B = \{b_1, b_2\}, C = \{c_1, c_2, c_3\}$. Let $f : C \rightarrow A$, $g : C \rightarrow B$ be embeddings that map $c_1, c_2, c_3$ to $\{a_1, a_2\}, \{a_3\}, \{a_4\}$, and $\{b_1\}, \{b_2\}, \emptyset$, respectively. Let $J \subseteq \Delta A$ be defined by the constraints $P(\{a_1, a_2\}) \geq 0.6$ and $P(\{a_4\}) = 0$. Let $H \subseteq \Delta B$ be defined by the constraint $P(\{b_1\}) \geq 0.6$. Then

$$\bar{f}(J) = \bar{g}(H) = \{P \in \Delta C \mid P(c_1) \geq 0.6, P(\{c_3\}) = 0\},$$

and $J = f(\bar{f}(J))$, $H = g(\bar{g}(H))$. Hence $J \xleftrightarrow{f,g} H$.

The partition $S_J$ is $\{\{a_1, a_2\}, \{a_3\}, \{a_4\}\}$; the partition $S_H$ is $\{\{b_1\}, \{b_2\}\}$.

Next, we compute $\tilde{I}_{me}(J)$ and $\tilde{I}_{me}(H)$. For $\tilde{I}_{me}(J)$ we first determine $J \upharpoonright S_J$, which is $\{P \in \Delta S_J \mid P(\{a_1, a_2\}) \geq 0.6, P(\{a_4\}) = 0\}$ (unlike here, it need not always be the case that the constraints defining $J \upharpoonright S_J$ are identical to the original constraints defining $J$). Thus, we get $I_{me}(J \upharpoonright S_J) = \{(0.6, 0.4, 0)\}$, and

$$\tilde{I}_{me}(J) = Ext(I_{me}(J \upharpoonright S_J))$$
$$= \{P \in \Delta A \mid P(\{a_1, a_2\}) = 0.6, P(a_4) = 0\}$$

Note that $I_{me}(J) = (\frac{1}{3}, \frac{1}{3}, \frac{1}{3}, 0) \notin \tilde{I}_{me}(J)$. Similarly, we compute $\tilde{I}_{me}(H) = \{P \in \Delta B \mid P(\{b_1\}) = 0.6\}$. Finally, we can check that

$$\bar{f}(\tilde{I}_{me}(J)) = \{P \in \Delta C \mid P(\{c_1\}) = 0.6, P(c_4) = 0\}$$
$$= \bar{g}(\tilde{I}_{me}(H)),$$

so that (9) is satisfied.

The proof of theorem 4.4, which was given in (Jaeger 1995), is not constructive. The following results show that at least in the case of $J$ being a polytope, $S_J$ can be effectively constructed.

**Theorem 4.8** Let $A = \{a_1, \ldots, a_n\}$, let $J \subseteq \Delta A$ be a polytope defined by $k$ linear inequality constraints $C_i = \sum_{j=1}^{n} r_{ij} P(a_j) \leq s_i$ $(i = 1, \ldots, k; r_{ij}, s_i \in \mathbf{R})$. Then the partition $S_J$ of $A$ can be computed in time polynomial in $kn$.

**Proof:** In the sequel, we denote $P(a_j)$ by $p_j$. Computation of $S_J$ amounts to deciding the equivalence relation

$$a_i \sim a_j :\leftrightarrow \{a_i, a_j\} \subseteq B \quad \text{for some } B \in S_J. \quad (12)$$

For notational convenience, take $i = 1, j = 2$. For $P = (p_1, p_2, \ldots, p_n) \in \Delta A$ we abbreviate $p_3, \ldots, p_n$ by **p**. We then get: $a_1 \sim a_2$ iff $\forall P = (p_1, p_2, \mathbf{p}) \in \Delta A$:

$$P \in J \leftrightarrow (p_1 + p_2, 0, \mathbf{p}) \in J \text{ and } (0, p_1 + p_2, \mathbf{p}) \in J. \quad (13)$$

Now define for $i = 1, \ldots, k$: $\bar{r}_i := max\{r_{i1}, r_{i2}\}$, and let $\bar{C}_i$ be the constraint defined by replacing $r_{i1}$ and $r_{i2}$ by $\bar{r}_i$ in $C_i$. Let $\bar{J}$ be the polytope defined by the $\bar{C}_i$. We show that

$$a_1 \sim a_2 \leftrightarrow J = \bar{J}. \quad (14)$$

The left to right direction in (14) is immediate: $J = \bar{J}$ implies that $J$ is definable by a set of constraints in which $p_1$ and $p_2$ only appear within terms of the form $\bar{r}(p_1 + p_2)$, so that (13) clearly is satisfied.

For the right to left direction, assume that $a_1 \sim a_2$. $\bar{J} \subseteq J$ trivially holds, because $\bar{J}$ is obtained from $J$ by sharpening each defining constraint. Let $P = (p_1, p_2, \mathbf{p}) \in J$. Consider $i \in \{1, \ldots, k\}$, and assume without loss of generality that $\bar{r}_i = r_{i1}$. By (13) we have that $(p_1 + p_2, 0, \mathbf{p}) \in J$. Since $(p_1 + p_2, 0, \mathbf{p})$ satisfies $C_i$ iff $(p_1, p_2, \mathbf{p})$ satisfies $\bar{C}_i$, we obtain that $P$ satisfies $\bar{C}_i$. This holds for all $i$, so that $J \subseteq \bar{J}$.

A test for $J \subseteq \bar{J}$ can be conducted via $k$ satisfiability tests for systems of $k + 1$ linear constraints on $n$ variables. Each such test can be conducted in time polynomial in $nk$ (e.g. Chvátal 1983). Finally, we have to do $\leq n(n-1)$ tests of the relation $\sim$, giving an overall runtime for the construction of $S_J$ polynomial in $nk$. $\square$

So far, we only have presented some limited empirical evidence that the selection functions $\tilde{I}$ can still be interesting and useful. Example 4.7 shows that with $\tilde{I}_{me}$ nontrivial inferences can be obtained. Also, if we apply $\tilde{I}_{me}$ to our example of section 4.1, we find that we now infer $P(A) = 0.5$, $P(A_1 \vee A_2) = 0.5$, and $P(B_1 \wedge B_2) = 0.5$, which arguably is the most reasonable result. Beyond the evidence provided by examples like these, the net result of definition 4.5 and theorem 4.6, so far, only is that we know representation independent selection functions to exist. This, in itself, is not exciting, because the trivial selection function $I$, with $I(J) = J$ for all $J$, already is representation independent. In order to evaluate the significance of theorem 4.6, we therefore have to look for general results on the properties of $\tilde{I}$. More specifically, the question of interest is: how many of the original useful properties of $I$ are preserved under the transformation $I \rightarrow \tilde{I}$, and what do we have to trade in for gaining representation independence?

The first major concession we have to make is precision: even where $I$ returns point values, $\tilde{I}$, in general, will only return intervals. By the impossibility result of Paris (1994) that was mentioned in section 4.1, this is an unavoidable weakness of representation independent selection functions. Another property that often holds for $I$, but usually is lost in $\tilde{I}$ is *continuity* (see Paris 1994) for a formal definition). To see why this is the case, consider the state space $A = \{a_1, a_2\}$, and $J_\epsilon \subseteq \Delta A$ defined by the constraint $P(a_1) \leq 1 - \epsilon$. For every $\epsilon > 0$ we then have $S_{J_\epsilon} = \{\{a_1\}, \{a_2\}\}$, and obtain, e.g., $\tilde{I}_{me}(J_\epsilon) = \{(0.5, 0.5)\}$. For $\epsilon = 0$, however, we have $J_0 = \Delta A$, $S_{J_0} = \{\{a_1, a_2\}\}$, and $\tilde{I}_{me}(J_0) = \Delta A$.

While the loss of precision and of continuity run counter conventional desiderata for a selection function, they both can be justified to some extent by arguing that the relevant state space we should consider is not so much the more or less arbitrarily specified underlying set $A$, as the set of "semantic concepts" implicit in our knowledge $J$. The elements of $S_J$ are just the formalization of such a notion of semantic concept, and the use of $\tilde{I}$ corresponds to using this set as the truly relevant state space. Introducing a constraint $P(a_1) \leq 1 - \epsilon$, then, more than establishing a numerical constraint, carries the impact of introducing $\{a_1\}$ as a semantic concept that we have to account for in our state space.

The following theorem provides some positive results on what is preserved under the transformation $I \rightarrow \tilde{I}$.

**Theorem 4.9** The principles of weak independence and irrelevant information are preserved under the transformation $I \rightarrow \tilde{I}$. If $I$ is dimension independent, then the relativization principle also is preserved under $I \rightarrow \tilde{I}$.

**Proof:** (Sketch) We only prove the statement for weak independence and irrelevant information, omitting the (simpler) proof for relativization.

The key to the proof is the following observation: if $J$ is of the form (3), then $S_J$ is essentially the product of $S_{J^B}$ and $S_{J^C}$. There is a minor complication: $S_{J^B}$ [$S_{J^C}$] may contain the element $B_0 := \{b \in B \mid \forall P \in J^B : P(b) = 0\}$ [the similarly defined $C_0$]. Then $S_J$ will contain the "irregular" set $A_0 := B \times C_0 \cup B_0 \times C$. The exact claim we want to prove is

$$S_J = (S_{J^B} \setminus B_0 \times S_{J^C} \setminus C_0) \cup A_0. \quad (15)$$

We denote the right hand side of (15) by $S^*$. It is straightforward to verify that $J = Ext(J \upharpoonright S^*)$, which shows that $S^*$ is a refinement of $S_J$. It remains to show, conversely, that $S^*$ is a refinement of $S_J$. This means that we have to show that there do not exist distinct elements $s_1^*, s_2^* \in S^*$ with $s_1^* \sim s_2^*$ (since $S^*$ is a refinement of $S_J$, the equivalence relation $\sim$ also is well-defined on $S^*$ by: $s_1^* \sim s_2^*$ iff $a_1 \sim a_2$ for some $a_1 \in s_1^*, a_2 \in s_2^*$). We proceed indirectly, and assume that such $s_1^*, s_2^*$ exist. The case where either $s_1^*$ or $s_2^*$ is equal to $A_0$ easily leads to a contradiction, so we assume that $s_1^* = (s_1^B, s_1^C)$, $s_2^* = (s_2^B, s_2^C)$ with $s_i^B \in S_{J^B} \setminus B_0$, $s_i^C \in S_{J^C} \setminus C_0$ $(i = 1, 2)$.

It follows that there exists $P \in J$ with $P(s_1^*) > 0$. From $s_1^* \sim s_2^*$, using criterion (13), it then follows that there also is $Q \in J$ with $Q(s_1^*) > 0$ and $Q(s_2^*) > 0$. Marginalizing on $B$ yields the result: there exists $Q^B \in J^B$ with $Q^B(s_1^B) > 0$ and $Q^B(s_2^B) > 0$. Let $\varepsilon = min\{Q^B(s_1^B), Q^B(s_2^B)\}$. From $s_1^C \not\sim s_2^C$ (with $\sim$ defined by $J^C$) it follows that there exists $Q_1^C \in J^C$, $Q_2^C \in \Delta C \setminus J^C$ that agree on all elements of $S_{J^C}$ except $s_1^C$ and $s_2^C$, and such that $|Q_1^C(s_1^C) - Q_2^C(s_1^C)| = \delta < \varepsilon$. Without loss of generality, assume that $Q_1^C(s_1^C)(= Q_2^C(s_2^C)) = Q_1^C(s_2^C) + \delta$. It is readily verified, that the two marginals $Q^B$ and $Q_1^C$ can be extended to a measure $O$ on $S^*$ with $O(s_1^*) = min\{Q^C(s_1^C), Q^B(s_1^B)\} \geq \delta$. From $s_1^* \sim s_2^*$ it follows that $O'$ obtained from $O$ by shifting probability mass $\delta$ from $s_1^*$ to $s_2^*$ again is in $J$. But now the marginal of $O'$ on $C$ is just $Q_2^C$, which thus would have to belong to $J^C$, contradicting our assumption. This completes the proof that $S^* \subseteq S_J$.

Having established (15), the remainder of the proof for weak independence and irrelevant information is simple. From the fact that $J \upharpoonright S_J$ is of the form (3), and the assumption that $I$ satisfies the respective principles, we obtain (2) and (5), respectively, for $\tilde{I}(J) \upharpoonright S_J = I(J \upharpoonright S_J)$. It is readily verified that the validity of (2) and (5) is preserved when then $I(J) \upharpoonright S_J$ is extended to $A$ via $Ext(I(J) \upharpoonright S_J)$.
□

The third line in table 1 summarizes the contents of theorem 4.9 and the negative results mentioned above. Line 4 and 5 also explicitly list the properties of $\tilde{I}_{me}$ and $\tilde{I}_{cm}$, some of which can be derived directly from the entries in lines 1-3; others require short, separate, proofs.

## 5    CONCLUSION

Rationality criteria for measure selection functions mostly have been formulated in terms of formal principles, or axioms. Entropy maximization has the best track record with respect to these principles; it is therefore often regarded as the one most reasonable selection rule. In this paper we have pointed out that rationality might also be based on the statistical criterion of identifying the most likely source for the information we are given, and that under such a changed perspective center of mass may look much more attractive than maximum entropy.

Representation independence is one intuitively reasonable formal principle that even entropy maximization fails to satisfy. It has been shown that we can gain representation independence when we are willing to forfeit some of the decisiveness of our inferences. This result is equally relevant for center of mass and for maximum entropy inference, as in particular it yields a language independent variant of center of mass. We have presented results that show that the modification of a selection rule $I$ to its representation independent variant $\tilde{I}$ preserves at least some of those characteristic features of $I$ that made $I$ attractive in the first place.

## References

Chvátal, V. (1983), *Linear programming*, W.H. Freeman and Co., New York, NY.

Druzdzel, M. J. & van der Gaag, L. C. (1995), Elicitation of probabilities for belief networks: Combining qualitative and quantitative information, *in* 'Proceedings of UAI-95', pp. 141–148.

Grove, A. J. & Halpern, J. Y. (1997), Probability update: Conditioning vs. cross-entropy, *in* 'Proceedings of UAI-97', pp. 208–214.

Halpern, J. Y. & Koller, D. (1995), Representation dependence in probabilistic inference, *in* 'Proceedings of IJCAI-95', pp. 1853–1860.

Jaeger, M. (1995), Default Reasoning about Probabilities, PhD thesis, Universität des Saarlandes.

Jaeger, M. (1996), Representation independence of nonmonotonic inference relations, *in* 'Proceedings of KR'96', pp. 461-472.

Paris, J. B. (1994), *The Uncertain Reasoner's Companion*, Cambridge University Press.

Paris, J. & Vencovská, A. (1990), 'A note on the inevitability of maximum entropy', *International Journal of Approximate Reasoning* **4**, 183–223.

Paris, J. & Vencovská, A. (1997), 'In defense of the maximum entropy inference process', *International Journal of Approximate Reasoning* **17**, 77–103.

Shore, J. & Johnson, R. (1980), 'Axiomatic derivation of the principle of maximum entropy and the principle of minimum cross-entropy', *IEEE Transactions on Information Theory* **IT-26**(1), 26–37.

Uffink, J. (1996), 'The constraint rule of the maximum entropy principle', *Studies in History and Philosophy of Modern Physics* **27**.